\documentclass[sigconf]{acmart}

\AtBeginDocument{%
  \providecommand\BibTeX{{%
    \normalfont B\kern-0.5em{\scshape i\kern-0.25em b}\kern-0.8em\TeX}}}

\setcopyright{rightsretained}
\copyrightyear{2021}
\acmYear{2021}
\acmDOI{}

\acmConference[Submitted to IMS '22]{Tel Aviv '22: 14th IFAC Workshop on Intelligent Manufacturing Systems}{March, 2022}{Tel Aviv, Israel}
\acmBooktitle{}
\acmPrice{}
\acmISBN{}

\usepackage{graphicx}      
\usepackage{natbib}        
\usepackage{listings}
\usepackage{xcolor}
\usepackage{adjustbox}
\usepackage{float}
\usepackage{color,soul}
\usepackage{aliascnt}
\usepackage{amsmath}
\usepackage{textgreek}
\usepackage{caption}
\usepackage{tikz-cd}
\usepackage{graphicx}
\usepackage{multirow}
\usepackage{hyperref}
\usepackage{academicons}
\usepackage{textcomp}
\restylefloat{table}

\begin{document}

\title{Streaming Machine Learning and Online Active Learning for Automated Visual Inspection.}


\author{Jo\v{z}e M. Ro\v{z}anec}
\authornotemark[1]
\orcid{0000-0002-3665-639X}
\affiliation{%
  \institution{Jo\v{z}ef Stefan International Postgraduate School}
  \streetaddress{Jamova 39}
  \city{Ljubljana}
  \country{Slovenia}}
\email{joze.rozanec@ijs.si}

\author{Elena Trajkova}
\orcid{0000-0001-5342-1085}
\affiliation{%
  \institution{University of Ljubljana, Faculty of Electrical Engineering}
  \streetaddress{Tr\v{z}a\v{s}ka 25}
  \city{Ljubljana}
  \country{Slovenia}
  \postcode{1000}
}
\email{trajkova.elena.00@gmail.com}

\author{Paulien Dam}
\orcid{0000-0001-5378-8100}
\affiliation{%
  \institution{Philips Consumer Lifestyle BV}
  \streetaddress{Oliemolenstraat 5}
  \city{Drachten}
  \country{The Netherlands}}
\email{paulien.dam@philips.com}

\author{Bla\v{z} Fortuna}
\orcid{0000-0002-8585-9388}
\affiliation{%
  \institution{Qlector d.o.o.}
  \streetaddress{Rov\v{s}nikova 7}
  \city{Ljubljana}
  \country{Slovenia}}
\email{blaz.fortuna@qlector.com}

\author{Dunja Mladeni\'{c}}
\orcid{0000-0003-4480-082X}
\affiliation{%
  \institution{Jo\v{z}ef Stefan Institute}
  \streetaddress{Jamova 39}
  \city{Ljubljana}
  \country{Slovenia}}
\email{dunja.mladenic@ijs.si}

\renewcommand{\shortauthors}{Trajkova and Ro\v{z}anec}

\begin{abstract}
Quality control is a key activity performed by manufacturing companies to verify product conformance to the requirements and specifications. Standardized quality control ensures that all the products are evaluated under the same criteria. The decreased cost of sensors and connectivity enabled an increasing digitalization of manufacturing and provided greater data availability. Such data availability has spurred the development of artificial intelligence models, which allow higher degrees of automation and reduced bias when inspecting the products. Furthermore, the increased speed of inspection reduces overall costs and time required for defect inspection. In this research, we compare five streaming machine learning algorithms applied to visual defect inspection with real-world data provided by \textit{Philips Consumer Lifestyle BV}. Furthermore, we compare them in a streaming active learning context, which reduces the data labeling effort in a real-world context. Our results show that active learning reduces the data labeling effort by almost 15\% on average for the worst case, while keeping an acceptable classification performance. The use of machine learning models for automated visual inspection are expected to speed up the quality inspection up to 40\%.
\end{abstract}




\maketitle

\section{Introduction}\label{S:INTRODUCTION}

\begin{figure*}[ht!]
\centering
\includegraphics[width=0.70\textwidth]{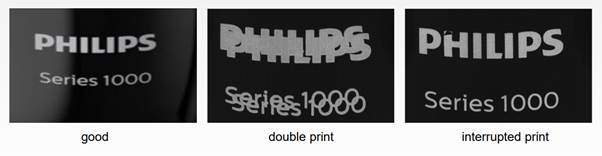}
\caption{Samples of each class in the dataset: good, double print, and interrupted print.}
\label{F:PHIA-CLASSES}
\end{figure*}

Quality control is a critical activity performed in manufacturing, which ensures products conform to specific requirements and specifications (\cite{yang2020using,kurniati2015quality,wuest2014approach}). Quality can be considered a competitive advantage and a business strategy, reducing rework in the production process, avoiding disruptions in the supply chain, and increasing consumer satisfaction (\cite{chan2003consumer,benbarrad2021intelligent}). Early defects detection enables tracing their root causes, take actions to mitigate them, enhancing the manufacturing process, and preventing the further manufacturing of defective parts.

While in many cases, quality inspection is visually performed by inspectors, such an approach has several drawbacks. First, the resources devoted towards training inspectors can grow proportionally to the scale of production. Second, the inspectors can work for a limited amount of time, and the quality of their work can be affected by the job design, environment, their experience, wellbeing, and motivation (\cite{cullinane2013job}). Third, due to the factors mentioned above, the quality of inspection can vary over time for a single inspector (\cite{kujawinska2016role}). Finally, the quality of inspection can be affected by the inspector-to-inspector consistency.

The decreased cost of sensors and connectivity has enabled an increasing digitalization of manufacturing (\cite{benbarrad2021intelligent}). The increasing democratization and adoption of Artificial Intelligence (AI) have been leveraged to enhance defects detection in manufacturing. Artificial Intelligence models provide virtually unlimited scalability capabilities, amend the drawbacks associated with the manual inspection (\cite{garvey2018framework,escobar2018machine,chouchene2020artificial}), and have been successfully applied to quality inspection in multiple scenarios (\cite{beltran2020external,napoletano2021semi,obregon2021rule,benbarrad2021intelligent}).


To realize supervised machine learning models for defect detection, data must be collected and annotated. While a first dataset is used to train a quality inspection model, increasing the number of labeled samples can enhance the performance of such a model. It is important to consider how informative the new samples are to the existing model. This principle is taken into account by active learning, a subfield of machine learning, which aims to identify which instances are most informative to the model and ask an \textit{oracle} to annotate them (\cite{settles2009active}). The \textit{oracle} can be a human annotator. The active learning strategy discards the non-informative samples.

We frame the defect detection problem as a supervised learning problem. In an optimal scenario, we consider that a production line would run without stopping (except for planned maintenance). Thus, the machine learning model should not require frequent deployments while delivering an acceptable defect discrimination performance. In addition, it is desired that manual labeling is minimized, reducing the manual effort and saving on data storage resources. We thus adopt a streaming active learning strategy and streaming machine learning algorithms. The streaming active learning strategy assumes a continuous stream of data, where for each instance, a decision must be made either to request its labeling or not. On the other hand, the streaming algorithm can issue a prediction on each new data instance and learn incrementally from those that are labeled by the oracle.

The main contributions of this research are (i) a comparative study between the five streaming machine learning algorithms for automated defect detection and (ii) two scenarios: no active learning (NO AL) and active learning (AL). We developed the machine learning models with images provided by the \textit{Philips Consumer Lifestyle BV} corporation. The dataset comprises shaver images classified into three categories, based on the defects observed regarding the logo printing: good, double print, and interrupted print (see Fig. \ref{F:PHIA-CLASSES}).

We evaluate the models using the area under the receiver operating characteristic curve (AUC ROC, see \cite{BRADLEY19971145}), a widely adopted classification metric. The metric is invariant to \textit{a priori} class probabilities, which makes it an excellent choice to measure classifiers performance on imbalanced datasets.

This paper is organized as follows. Section \ref{S:RELATED-WORK} describes related works, Section \ref{S:USE-CASE} describes the \textit{Philips Consumer Lifestyle BV} use case, Section \ref{S:METHODOLOGY} provides a brief description of the methodology and experiments we conducted, while Section \ref{S:RESULTS-AND-ANALYSIS} outlines the results obtained. Finally, Section \ref{S:CONCLUSION} concludes and describes future work.

\section{Related Work}\label{S:RELATED-WORK}
Automated visual inspection refers to image processing techniques used to perform quality control (\cite{beltran2020external}). Automated visual inspection techniques can be divided into two groups (supervised and unsupervised methods), depending on whether labeled or unlabeled data is required. Supervised machine learning methods require a dataset of labeled images, from which features are extracted to train the machine learning model later. Current state-of-the-art image processing techniques frequently make use of deep learning models, whether it is for end-to-end learning, or as feature extractors on top of which other models can be developed (\cite{long2015fully,glasmachers2017limits,pouyanfar2018survey}), though manually engineered features are used too (\cite{ravikumar2011machine}).

The adoption of automated visual inspection for defect detection is being fostered in the context of Industry 4.0 (\cite{carvajal2019online}). While it was applied in several use cases in the past, it is considered that the field is still in its early stages and that artificial intelligence can revolutionize product inspection (\cite{aggour2019artificial}). Machine learning for visual inspection has been applied to control the quality of printing (\cite{villalba2019deep}), inspect printed circuit board production (\cite{duan2012machine}), determine whether bottles met the required specifications (\cite{sahoo2019dynamic}), detect defects during metallic powder bed fusion in additive manufacturing (\cite{gobert2018application}), manufactured vehicle parts (\cite{liqun2020research}), aerospace components (\cite{beltran2020external}), or textile quality control (\cite{jiang2018fundamentals}). While the aforementioned authors report using machine learning algorithms, such as the Support Vector Machines, k-Nearest Neighbours, or Multilayer Perceptrons, in a batch setting, \cite{lughofer2017line} describes a visual inspection use case where evolving fuzzy classifiers are used. In addition, an online (streaming) active learning setting is used to reduce manual labeling work.

Active learning (AL) is a sub-field of machine learning that assumes that many unlabeled data exists while the cost of labeling those data instances is expensive. Thus, it explores different techniques to find and select the most informative data instances, which are then presented to an \textit{oracle} (e.g., a human annotator), who can label them. Online active learning assumes a stream of unlabeled instances exists, where one data instance is drawn at a time, and a decision must be made whether to label or discard such piece of data (\cite{settles2009active}). One such criterion is uncertainty sampling, where the sample is selected based on the measured uncertainty according to a particular metric or machine-learning model (\cite{lewis1994heterogeneous}). While active learning was applied in the manufacturing domain, not much research has been performed in this direction (\cite{meng2020machine}).

Streaming machine learning is a sub-field of machine learning that studies how a learner can provide adequate forecasts while learning from a sequence of data instances, learning one data instance at a time (\cite{hoi2018online}). Such models can be immediately updated and thus provide high scalability when dealing with large amounts of data arriving at a high velocity (\cite{gomes2019machine}). In this work we compare five streaming classification algorithms: Hoeffding Tree (HT) (\cite{hulten2001mining}), Hoeffding Adaptive Tree (HAT) (\cite{bifet2009adaptive}), Stochastic Gradient Tree (SGT) (\cite{gouk2019stochastic}), streaming Logistic Regression (SLGR), and streaming k-Nearest Neighbors (SKNN).

Combining online active learning and streaming machine learning classifiers allows to reduce downtimes required for new model deployments, updating the classification model one instance a time, and enabling an \textit{inspection by exception}, only requesting the labeling of images for which ones the model is most uncertain about (\cite{chouchene2020artificial}).

\section{Use Case}\label{S:USE-CASE}

\begin{figure*}[ht]
\centering
\includegraphics[width=15cm]{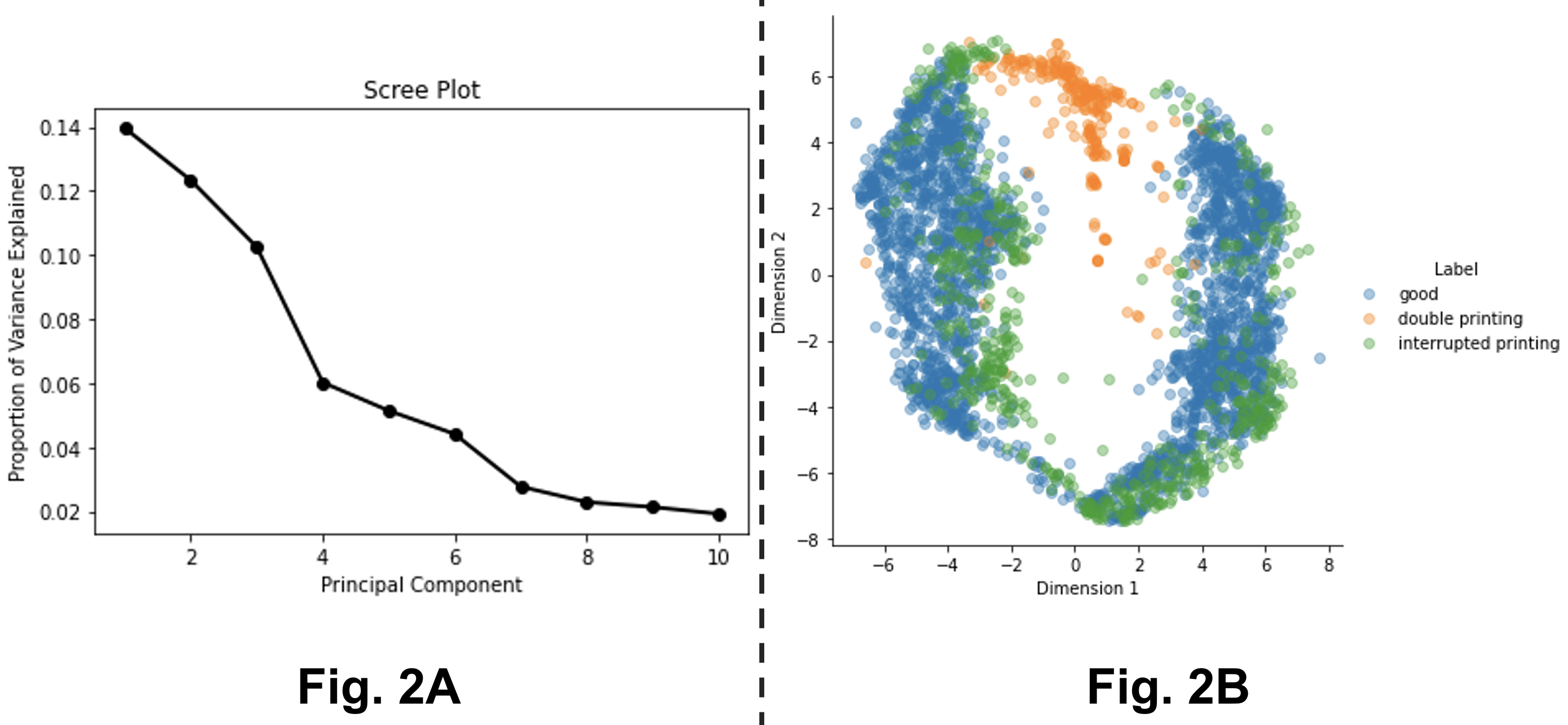}
\caption{Fig. \ref{F:TSNE}A shows a scree plot used to assess dimensionality reduction before creating the t-SNE plot. While for the t-SNE plot, we use ten variables, these could be further reduced to eight. Fig. \ref{F:TSNE}B shows a t-SNE plot, with clusters of data points that correspond to the three target labels.}
\label{F:TSNE}
\end{figure*}

In this research, we focus on the visual inspection task performed on shavers produced by \textit{Philips Consumer Lifestyle BV}. Such inspection aims to determine if the \textit{Philips Consumer Lifestyle BV} company logo was properly printed and eventually detect the kind of defective printing. To that end, the company uses different pad-printing setups. A vast amount of products are produced every day, each of which is manually handled and inspected. If a defective print is found, the product is removed from the production line. It is expected that the use of an automated visual quality inspection system would strongly reduce manual work, speeding up the process by more than 40\%. A labeled dataset of 3.518 images was provided for the task at hand to train and test the models. Two kinds of defects were labeled (double printing and interrupted printing), in addition to good prints. We treat the problem as a multiclass classification task. The images distribution can be visualized in the t-SNE plot at Fig. \ref{F:TSNE}. A scree plot was used to assess the best amount of variables to consider when performing the dimensionality reduction.

\section{Methodology}\label{S:METHODOLOGY}

We framed the automated defect detection as a multiclass imbalanced classification problem. We make use of a pre-trained ResNet-18 model (\cite{he2016deep}) as a feature extractor to obtain image embeddings. In particular, we extract features from the Average Pooling Layer, which translates each image into a feature vector of 512 values. To prevent overfitting, we perform feature selection selecting the \textit{top K} features ranked by mutual information (\cite{kraskov2004estimating}), considering $K=\sqrt{N}$, where N is the number of data instances in the train set (\cite{hua2005optimal}).

To compare the performance of different streaming machine learning algorithms and assess the benefits of a streaming active learning approach, we divided the dataset using a stratified ten-fold cross-validation (\cite{zeng2000distribution,kuhn2013applied}), using one fold for testing, and divided the remaining data into the training set, and a stream of data for the active learning setting, simulating unlabeled samples that could be labeled on request. We always queried the unlabeled instances until exhausting the stream. When doing so, we asked for labels of those unlabeled instances with a classification uncertainty above 0.55. Then, we measured the models ' performance over the test set for each new instance obtained from the unlabeled data stream. To that end, we computed the AUC ROC metric with the "one-vs-rest" heuristic method. The "one-vs-rest" heuristic splits the multiclass dataset into multiple binary classification problems, computing the AUC ROC metric for each class and summarizing the final value as the weighted average considering the number of true instances for each class. We repeated the procedure mentioned above ten times to estimate better the performance of the models and learning scenarios (no active learning, and with active learning).

\begin{figure*}[ht]
\centering
\includegraphics[width=16cm]{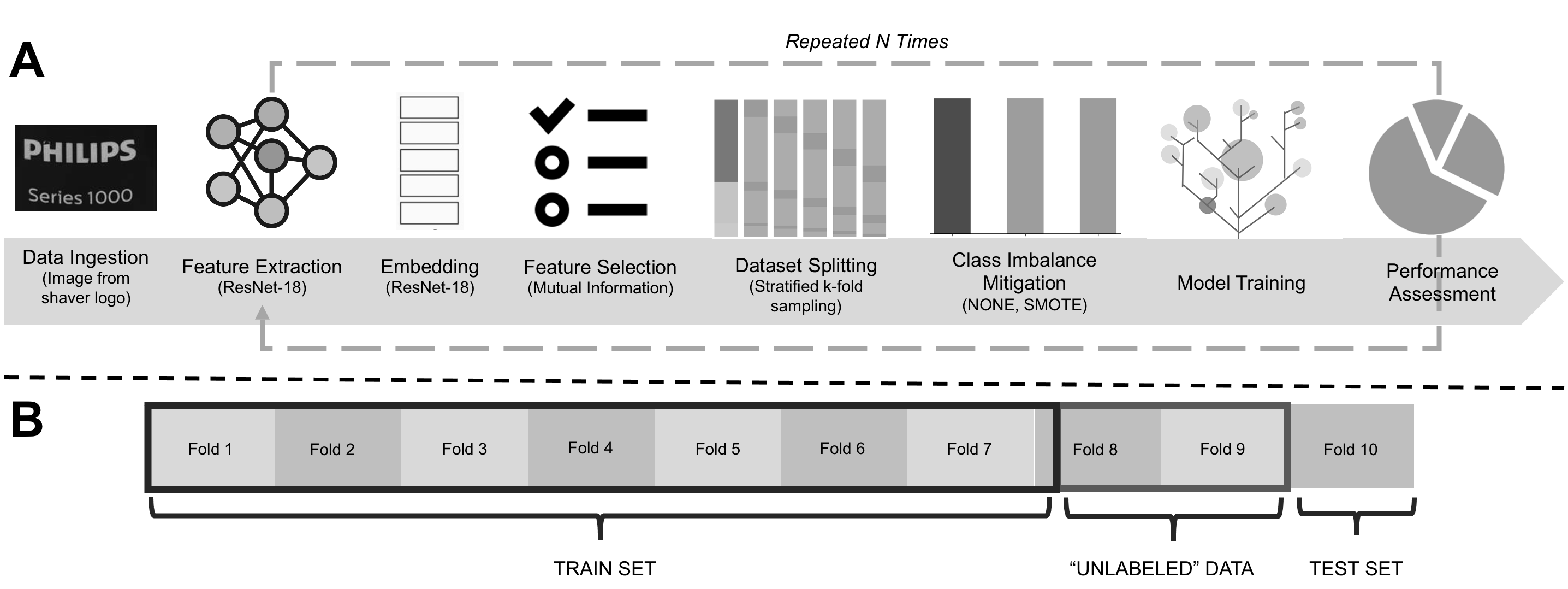}
\caption{Fig. \ref{F:METHODOLOGY}A shows the methodology we followed to develop and assess the models. In Fig. \ref{F:METHODOLOGY}B we depict how data was partitioned into train, test, and a stream of instances used to simulate unlabeled data and their labeling in the active learning scenario.}
\label{F:METHODOLOGY}
\end{figure*}


\section{Results and Analysis}\label{S:RESULTS-AND-ANALYSIS}

\begin{table*}[ht!]
\centering
\resizebox{0.80\textwidth}{!}{
\begin{tabular}{|l|r|r|r|}
\hline
\textbf{SCENARIO} & \multicolumn{1}{l|}{\textbf{AUC ROC\textsubscript{MEAN+CI}}} & \multicolumn{1}{l|}{\textbf{AUC ROC\textsubscript{MEAN+CI Q1}}} & \multicolumn{1}{l|}{\textbf{AUC ROC\textsubscript{MEAN+CI Q4}}} \\ \hline
\textbf{NO AL} & \textbf{0,6350\textpm0,0007} & \textbf{0,6085\textpm0,0013} & \textbf{0,6532\textpm0,0015} \\ \hline
\textbf{AL} & \textit{0,6025\textpm0,0007} & \textit{0,5903\textpm0,0012} & \textit{0,6103\textpm0,0015} \\ \hline
\end{tabular}
\caption{The table describes average AUC ROC values, and confidence intervals (CI) at a confidence level of 95\%, obtained across ten times repeated ten-fold cross-validation for two scenarios. Best results are bolded, second-best results are highlighted in italics. Q1 and Q4 denote the first and last quartiles, respectively. \label{T:AUC-ROC-VALUES-SCENARIOS}}}
\end{table*}

\begin{table*}[ht!]
\centering
\resizebox{0.45\textwidth}{!}{
\begin{tabular}{|l|r|r|r|}
\hline
\textbf{MODEL} & \multicolumn{1}{l|}{\textbf{NO AL}} & \multicolumn{1}{l|}{\textbf{AL}} \\ \hline
SLGR & \textit{0,7655\textpm0,0010} & \textit{0,6920\textpm0,0012} \\ \hline
SKNN & \textbf{0,8266\textpm0,0005} & \textbf{0,8186\textpm0,0004} \\ \hline
SGD & 0,5194\textpm0,0005 & 0,5144\textpm0,0004 \\ \hline
HT & 0,5164\textpm0,0005 & 0,4972\textpm0,0004 \\ \hline
HAT & 0,5470\textpm0,0010 & 0,4901\textpm0,0005 \\ \hline
\end{tabular}
\caption{The table presents mean AUC ROC values and confidence intervals (CI) at a confidence level of 95\%, obtained for five streaming machine learning algorithms across two scenarios when performing a ten times repeated ten-fold cross-validation. Best results are bolded, second-best results are highlighted in italics. \label{T:AUC-ROC-VALUES-MODELS}}}
\end{table*}

\begin{table*}[ht!]
\centering
\resizebox{0.75\textwidth}{!}{
\begin{tabular}{|l|r|r|r|r|r|r|r|}
\hline
\textbf{MODEL} & \multicolumn{1}{l|}{\textbf{MIN}} & \multicolumn{1}{l|}{\textbf{MAX}} & \multicolumn{1}{l|}{\textbf{MEAN}} & \multicolumn{1}{l|}{\textbf{STD DEV}} & \multicolumn{1}{l|}{\textbf{Q1}} & \multicolumn{1}{l|}{\textbf{Q2}} & \multicolumn{1}{l|}{\textbf{Q3}} \\ \hline
SLGR & \textit{13,21\%} & \textbf{100,00\%} & 35,11\% & 32,50\% & \textit{17,76\%} & \textit{19,89\%} & 21,70\% \\ \hline
SKNN & 0,71\% & \textit{68,61\%} & 14,72\% & 25,46\% & 1,70\% & 2,13\% & 3,08\% \\ \hline
SGD & 0,00\% & \textbf{100,00\%} & 20,00\% & 40,00\% & 0,00\% & 0,00\% & 0,00\% \\ \hline
HT & 0,00\% & \textbf{100,00\%} & \textit{39,42\%} & 44,45\% & 0,00\% & 0,00\% & \textit{86,79\%} \\ \hline
HAT & \textbf{17,76\%} & \textbf{100,00\%} & \textbf{75,12\%} & 25,40\% & \textbf{52,45\%} & \textbf{92,04\%} & \textbf{96,98\%} \\ \hline
\end{tabular}
\caption{The table quantifies how much labeling effort is saved in a streaming active learning setting, considering models developed for different streaming machine learning algorithms. Only data instances with a classifier uncertainty higher than 0.55 were presented to the user. Measurements are taken for ten-fold cross-validation, repeated ten times. Best results are bolded, second-best results are highlighted in italics. Q1, Q2, and Q3 denote quartiles one to three, respectively. \label{T:GAINS}}}
\end{table*}

We compared two scenarios by running five streaming machine learning classification algorithms across them. The scenarios of interest were: streaming without active learning (NO AL) and streaming active learning labeling only data instances with an uncertainty higher than 0.55 (AL). We summarize the results obtained for both scenarios in Table. \ref{T:AUC-ROC-VALUES-SCENARIOS}. We found that the best mean discrimination was achieved without using active learning (NO AL): active learning degraded the models' discrimination performance. In both scenarios, we observed that the models' mean performance improved over time.

We present the models' performance across both scenarios in Table \ref{T:AUC-ROC-VALUES-MODELS}. The results show that the SKNN model achieved the best discrimination performance in both cases. The second-best results were reported for the SLGR. The worst performance was achieved by the HF and HAT models, which had a higher false-positive rate than the true-positive rate in the \textit{AL} scenario.

To understand the contribution of the active learning approach regarding manual labeling, we measured how much effort was saved (effort gain) as the percentage of all non-labeled samples that were not presented to the \textit{oracle} (human annotator) to request a label. We present the results in Table \ref{T:GAINS}. We found that all the algorithms reported some effort gain on average. The HAT reported the highest gains, which skipped a median of 92\% of the unlabeled samples, and reported effort gains of almost 97\% at the third quartile. We consider the second-best algorithm, in terms of effort gain, was the SLGR, which did not achieve higher mean savings than the HT, but saved some labeling effort in almost every case, which is reflected by the values observed at all quartiles. We consider the best overall performance (when measuring effort gain and discriminative power assessed through the AUC ROC), was obtained by the SLGR algorithm, which had the second best performance in terms of discriminative power, and competitive results in terms of effort gain. 

\section{Conclusion}\label{S:CONCLUSION}
This paper compared five streaming machine learning algorithms across two scenarios: streaming without active learning (NO AL) and streaming with a streaming active learning setting (AL). Our results show that all algorithms improve their performance over time regardless of active learning is used. We consider the best overall performance was achieved by the SLGR algorithm, which achieved the second-best scores regarding discriminative power while also reporting competitive effort gains in terms of manual labeling effort (reduced 35\% on average). While the SKNN algorithm obtained better results in terms of discriminative power, it reported low gains in manual labeling effort (nearly 15\% on average). We thus conclude that active learning can provide effective means towards reducing the manual labeling effort. However, special attention must be put to the model to understand the relationship between models' learning, discriminative performance, and effective manual labeling effort gains. Regardless of the active learning setting, experts estimate that the deployment of machine learning defect detection models can speed up visual inspection up to 40\% for this particular use case. Future work will study different approaches to tackle class imbalance in a streaming setting and further improve the classification performance. In addition, we will explore methods to provide cues to the human annotator, showing whether an error is expected and where it could be located. We expect such a piece of information will help to speed up the annotation times. Furthermore, we will develop explainable artificial intelligence methods to understand which segments of the image are used by the classifier, assess the quality of their learning, and identify opportunities for further model improvement.

\begin{acks}
This work was supported by the Slovenian Research Agency and the European Union’s Horizon 2020 program project STAR under grant agreement number H2020-956573. The authors acknowledge the valuable input and help of Jelle Keizer and Yvo van Vegten from \textit{Philips Consumer Lifestyle BV}.
\end{acks}

\bibliographystyle{ACM-Reference-Format}
\bibliography{main}

\end{document}